\documentclass{llncs}

\usepackage{graphicx}
\usepackage[bookmarks=false]{hyperref}

\usepackage[acronym]{glossaries}
\newacronym{EC}{EC}{Evolutionary Computation}
\newacronym{IEC}{IEC}{Interactive Evolutionary Computation}
\newacronym{GA}{GA}{Genetic Algorithm}
\newacronym{IGA}{IGA}{Interactive Genetic Algorithm}
\newacronym{RMSE}{RMSE}{Root Mean Square Error}
\newacronym{ML}{ML}{Machine Learning}
\newacronym{DCNN}{DCNN}{Deep Convolutional Neural Network}
\newacronym{DL}{DL}{Deep Learning}
\newacronym{DNN}{DNN}{Deep Neural Network}
\newacronym{CNN}{CNN}{Convolutional Neural Network}
\newacronym{OCR}{OCR}{Optical Character Recognition}

\newcommand{\urlVideoDemo}{\texttt{\href{https://cdv.dei.uc.pt/2018/evostar/evotype_demo.mov}{cdv.dei.uc.pt/2018/evostar/evotype\_demo.mov}}}
\newcommand{\urlVideoResults}{\texttt{\href{https://cdv.dei.uc.pt/2018/evostar/evotype_results.mov}{cdv.dei.uc.pt/2018/evostar/evotype\_results.mov}}}

\begin{document}

\title{Evotype: Towards the Evolution of Type Stencils}

\author{Tiago Martins\and Jo\~{a}o Correia\and Ernesto Costa\and Penousal Machado}
\authorrunning{Martins et al.}
\institute{CISUC, Department of Informatics Engineering,\\%
University of Coimbra, 3030 Coimbra, Portugal\\%
\texttt{\{tiagofm,jncor,ernesto,machado\}@dei.uc.pt}}
\toctitle{Evotype: Evolutionary Type Design}
\tocauthor{Martins et al.}

\maketitle

\begin{abstract}
Typefaces are an essential resource employed by graphic designers. The increasing demand for innovative type design work increases the need for good technological means to assist the designer in the creation of a typeface. We present an evolutionary approach for the automatic generation of type stencils to draw coherent glyphs for different characters. The proposed system employs a Genetic Algorithm to evolve populations of type stencils. The evaluation of each candidate stencil uses a hill climbing algorithm to search the best configurations to draw the target glyphs. We study the interplay between legibility, coherence and expressiveness, and show how our framework can be used in practice.
\end{abstract}


\section{Introduction}\label{sec_introduction}

Typefaces are an essential resource employed by graphic designers \cite{lupton2004a}. Most designers choose typefaces from vast libraries of existing examples. However, some opt to create their own typefaces or custom lettering that better suits the design problem. This need to create and customize typefaces is, perhaps, due to the fact that there is no universal typeface that fits every design project. Each project is unique. For this reason, the selection of the typeface(s) for one project should take into account its specifications and requirements. Also, there will always be designers willing to create and experiment with type, and there will always be people pursuing novelty.

Type design is a laborious process in which the designer has to create several coherent glyphs to form a typeface. A glyph consists in a concrete design of a character, \emph{e.g.}, an individual letter, figure, or punctuation mark. This, along with the increasing demand for innovative type design work, increases the need for good technological means to assist the designer in the creation of a typeface.

In this paper, we continue our research on automatic evolution of glyphs \cite{martins2015a,martins2016a}. We explore a different approach based on type stencils (see, \emph{e.g.}, figure~\ref{fig_pdu}). It employs \gls{EC} techniques to automatically create stencils to draw glyphs for different characters. Each evolving stencil consists of an arrangement of visual elements, such as line segments, that make up the glyphs. The recipe that defines how the stencil can be used to draw each glyph is also evolved.

With our system, we are able to achieve new forms of typographic expression while maintaining unity and coherence across the generated glyphs. The expressiveness of the glyphs can be explored by transforming and replacing the stencils' elements with custom shapes and strokes that embody the glyphs according to the system defined by the stencil.

The main contribution presented herein is a generative system capable of automatically creating stencils to draw coherent glyphs. Other contributions include: %
(i) an exploration of how evolutionary approaches may inform contemporary graphic design practices; %
(ii) an investigation into the interplay between legibility, coherence, and expressiveness; %
(iii) a generic evolutionary framework that integrates an evolutionary and an evaluation module for the generation of stencils to draw coherent glyphs with different degrees of detail and abstraction; %
(iv) an investigation into the relationships among the different parts that build glyphs; %
and (v) a showcase of the application of the generated results, allowing designers to create new type explorations.

The remainder of this paper is organised as follows: section~\ref{sec_related_work} summarises the related work focusing on applications of evolutionary techniques in type design; section~\ref{sec_system} overviews the behaviour of the proposed system; section~\ref{sec_experiments} describes a series of experiments conducted to explore and analyse the possibilities created by the system, and presents the results for each experiment; finally, section~\ref{sec_conclusions} presents conclusions and directions for future work.

\section{Related Work}\label{sec_related_work}

\gls{EC} has been used in creative domains for the exploration of innovative solutions with success (see, \emph{e.g.}, \cite{mrm07a}). Nevertheless, as far as we know, only a few evolutionary approaches for type design exist. %
Butterfield and Lewis \cite{butterfield2000a} employ \gls{IEC} to present populations of deformed letters. Surface primitives encoded in the genotype of each candidate individual allow the deformation of the letters of a given typeface. %
Lund \cite{lund2000a} uses \gls{IEC} to evolve the settings for a parametric typeface. Each parameter controls a particular visual characteristic of the typeface. %
Levin et al. \cite{levin2001a} present the interactive system \emph{Alphabet Synthesis Machine} that allows the user to generate abstract letterforms. The system employs a \gls{GA} to evolve a population of letterforms according to fitness metrics obtained from an initial seed glyph provided by the user. %
Unemi and Soda \cite{unemi2003a} developed an \gls{IEC}-based system for the design of Japanese Katakana glyphs. It allows the construction of glyphs using simple elements that are controlled by parameters encoded in the genotype and drawn along a predefined skeleton. %
Schmitz \cite{schmitz2004a} presents the interactive program \emph{genoTyp}, where the typefaces are generated according to genetic rules. The program allows the user to experiment with the breeding of existing typefaces as well as the manipulation of their genes, \emph{i.e.}, vertexes. %
Kuzma \cite{kuzma2008a} investigates the potential of \gls{IEC} by implementing the \emph{Font Evolving System}, which allows the user to evolve fonts interactively. %
Yoshida et al. \cite{yoshida2010a} present the \emph{Personal Adapted Letter}. In this approach, the user starts by manually defining parts of the glyphs. It employs an \gls{IGA} to modify the parts, presenting different options for the user to choose from in each iteration. %
In \cite{martins2015a}, we present \emph{Evotype}, which employs a \gls{GA} to evolve glyphs for the Roman alphabet using line segments. This approach uses an automatic fitness assignment scheme to guide the evolutionary process. In \cite{martins2016a}, we use a \gls{GA} to create glyphs using assemblages of shapes. The system uses different automatic fitness assignment schemes designed to explore the expressiveness and legibility of the generated glyphs.

The existing systems that apply evolutionary techniques for type design have some limitations and drawbacks. In most of them, the user is asked to manually guide the evolutionary process by selecting the ideal solutions in each generation until getting an acceptable one, \emph{i.e.}, they rely on user evaluation. Therefore, these approaches suffer from the well-known limitations of \gls{IEC}, namely the user fatigue, the consequent loss of interest, and inconsistent evaluation. %
Additionally, some approaches require pre-existing typefaces (\emph{e.g.}, \cite{butterfield2000a,schmitz2004a,kuzma2008a}) or skeletons (\emph{e.g.}, \cite{unemi2003a}). Others require the drawing of initial seed glyphs (\emph{e.g.}, \cite{levin2001a}), or the identification of letter parts (\emph{e.g.}, \cite{yoshida2010a}). Some need the creation of parametric typefaces (\emph{e.g.}, \cite{lund2000a}). Finally, parametrised approaches can overly influence their outcome. We observe the same issues in approaches that provide a limited range of visual elements to create the glyphs (\emph{e.g.}, \cite{martins2015a}). Others lack coherence between the glyphs of the same typeface (\emph{e.g.}, \cite{martins2016a}).


From our analysis of the related and previous work, we believe that: fitness assignment should be automatic, and the system should aid the designers by providing glyphs that can be further refined. We also consider important the visual coherence between glyphs, \emph{i.e.}, the presence of common visual characteristics between them. These two points guided the research and development of this work.

\section{System Overview}\label{sec_system}

\begin{figure}[!h]
    \centering
    \includegraphics[width=0.59\textwidth]{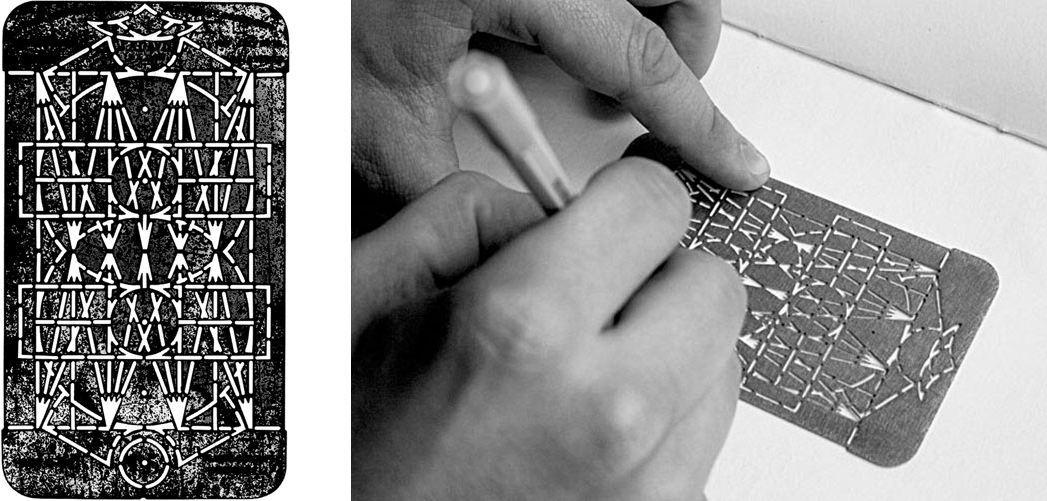}
    \caption{\emph{Plaque D\'{e}coup\'{e}e Universelle (PDU)}, Joseph A. David, 1876}
    \label{fig_pdu}
\end{figure}

This work is based on the idea of a stencil capable of generating every letter of the alphabet in a coherent manner. In 1876, the American engineer Joseph A. David developed the \emph{Plaque D\'{e}coup\'{e}e Universelle}, mostly known as the PDU (see figure~\ref{fig_pdu}). This stencil consists of a complex grid that allows the construction of all uppercase and lowercase letters, numbers, punctuation, accents, etc. \cite{kindel2007a}. A few decades later, the seven-segment display is invented. It employs a similar approach as the PDU by lighting its segments in different combinations to represent figures and letters.

A stencil can be seen as a system of rules to construct the glyphs. Similarly, the design of typefaces typically involves the creation of a set of parts or modules that the designer combines and reuses to form the different glyphs of a typeface. This way, most glyphs consist of a combination of smaller parts (see, \emph{e.g.}, \cite{craig2012a} for more details about type anatomy). One can analyse existing typefaces and observe the repetitive usage of these parts among glyphs, providing coherence between them.

With the idea of a unifying grid as the basis for this work, we developed a system that evolves type stencils and thus explore the interplay between the flexibility and the restrictions they provide while designing glyphs.

\begin{figure}[!b]
    \centering
    \includegraphics[width=0.85\textwidth]{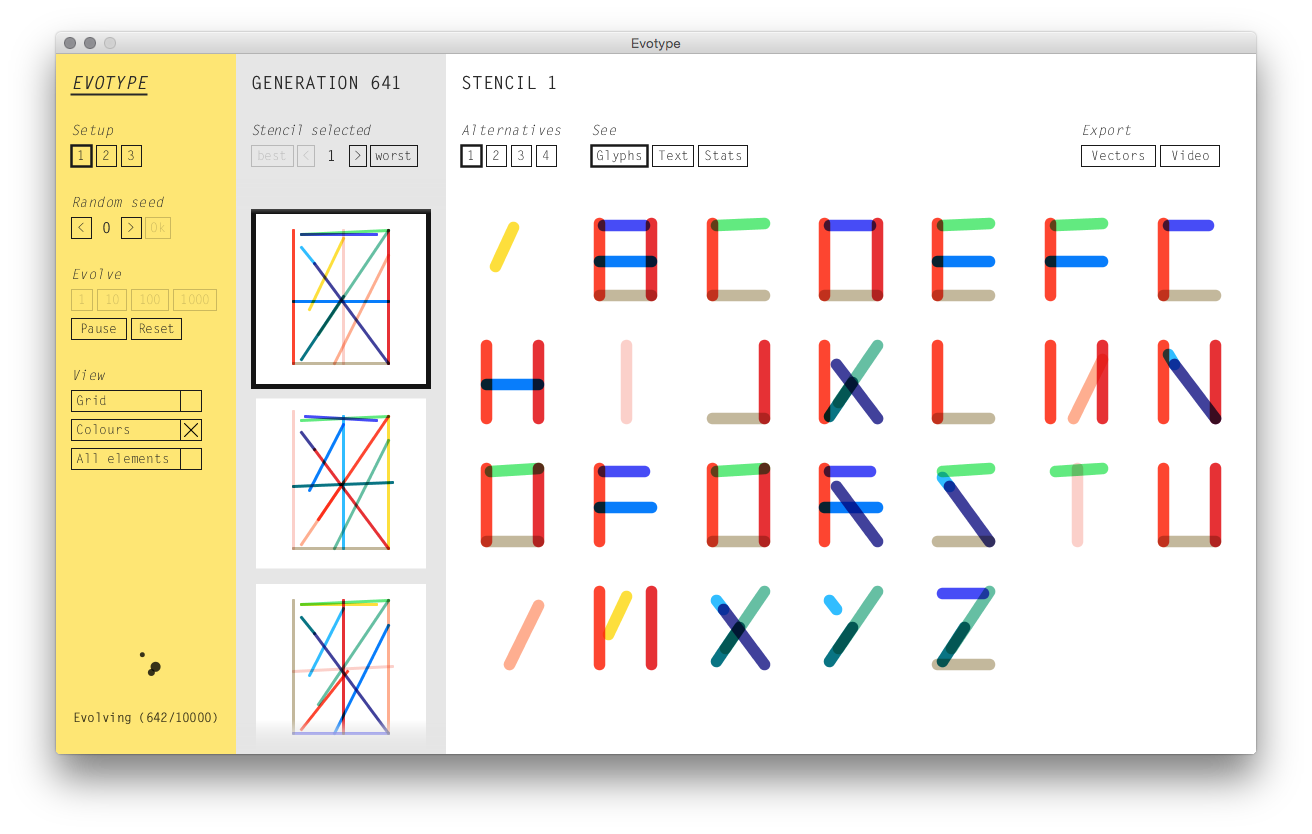}
    \caption{Screen shot of the system. A demo video can be seen at \urlVideoDemo.}
    \label{fig_screen_shot}
\end{figure}

Figure~\ref{fig_screen_shot} shows a screen shot of the system. A simple graphic user interface was developed to provide the user with the necessary means to evolve and export type stencils. While stencils are being evolved, the user can visualise and inspect the results in different ways: %
(i) browse throughout all candidate stencils of the current generation, which are arranged vertically by descending order of fitness; %
(ii) visualise for each character the different alternative glyphs drawn with the selected stencil; %
(iii) experiment with the glyphs of the selected stencil by writing a text with them and this way assess readability; %
and (iv) visualise a series of measurements that characterise the results in different aspects, \emph{e.g.}, fitness of the stencils, fitness of the glyphs, and the complexity of the selected stencil. Furthermore, the user can export any evolved stencil to make further refinements and use it. A demo video of the system can be seen at \urlVideoDemo.

The system integrates two main modules: (i) the evolution module, which implements a \gls{GA} to generate candidate stencils (see subsection~\ref{subsec_evolution}), and (ii) the evaluation module, which implements automatic fitness assignment schemes to evaluate the glyphs provided by the first module (see subsection~\ref{subsec_evaluation}).

\subsection{Evolution}\label{subsec_evolution}

The system employs a \gls{GA} to evolve a population of stencils. A \gls{GA} improves a set of candidate solutions, which in the case of this work are stencils, by iteratively employing methods of selection of the most promising for reproduction with variation. See, \emph{e.g.}, \cite{eiben2015a} for more details about \gls{GA}s.

\subsubsection{Representation}\label{subsubsec_representation}

Each stencil being evolved with the system is constructed from line segments. Therefore, each gene that composes the genotype of each stencil encodes one line segment in a two-dimensional space. Each gene consists of a 4-tuple with values corresponding to the coordinates of the end points of the line segment: (\texttt{x1}, \texttt{y1}, \texttt{x2}, \texttt{y2}). The position of each end point is constrained by a square grid with a given density, \emph{i.e.}, the end points adhere to the grid points. Different stencils may have a different number of line segments. Therefore, the size of the genotype, \emph{i.e.}, number of genes, may vary from stencil to stencil.

\begin{figure}
    \centering
    \includegraphics[width=0.66\textwidth]{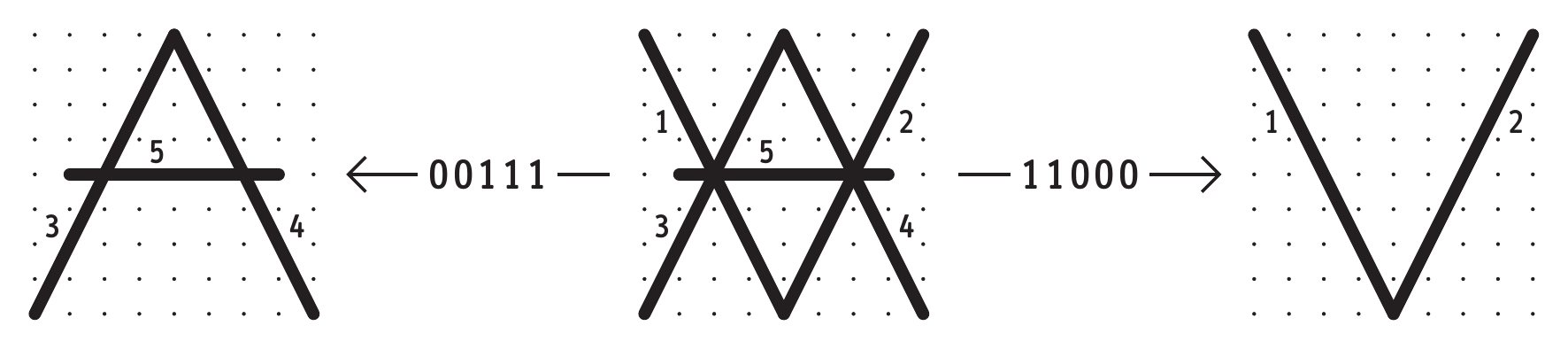}
    \caption{Mapping mechanism expressing the genotype of a stencil into two glyphs}
    \label{fig_mapping}
\end{figure}

The mapping mechanism that expresses each genotype into its perceptible artefact, or phenotype, consists in the drawing of black line segments encoded in the genotype on a white canvas. However, in this work, the mapping process is not direct. One mapping for each character we want to draw with the stencil is required. This way, we developed a mapping mechanism based on binary masks that define how a given stencil is used to draw a given glyph. When we say \emph{how}, we mean which line segments are used. This mechanism is schematically represented in figure~\ref{fig_mapping}.

\subsubsection{Variation Operators}\label{subsubsec_variation}

The initial population is seeded with randomly created stencils. During the evolutionary process, new stencils are created by applying variation operators such as crossover and mutation.

The crossover operator exchanges line segments between two stencils. We performed some preliminary experiments to study the impact of different crossover operators. 
We use a custom crossover operator that we have already tested in previous work \cite{martins2015a}, which uses a randomly selected area at the phenotype to determine the genes that are exchanged between the parents. 

Analysing the results of these experiments, we observed that the one-point crossover produced individuals with similar elements positioned in the same areas of the phenotypes. This was not observed using the second crossover operator, which avoided this accumulation of line elements in the phenotype. This is perhaps because the crossover operator exchanges line segments between stencils according to their position on the phenotype regardless of their position in the genotype. Note that the order of the genes in the genotype is irrelevant when expressing it into the phenotype. This way, we concluded that the area crossover operator performs better in this problem and for this reason, we adopted it in our system.

The crossover operator that we adopt proceeds as follows: (i) selects a random rectangular area of the phenotype; (ii) determines, for each parent stencil, the line elements whose middle points are inside the selected area; and (iii) exchanges those line elements between the parents. This crossover may be asymmetric as the number of genes it moves from the individual A to individual B may be different of from the number of genes it moves from B to A. This results in stencils with a different number of elements in comparison with their parents.

Mutating an evolving stencil involves random modification of some parts of its genotype. The mutation operator comprises three procedures: deletion, modification, and insertion of genes. Each can occur independently with preset probabilities. The deletion procedure selects a gene at random and removes it from the genotype. The modification procedure changes one of the end points of one or more line segments by moving it by the minimum translation in the grid in one of the eight possible directions. Finally, the insertion procedure inserts a new randomly generated gene into the genotype. The second procedure (modification) provides high locality, \emph{i.e.}, mutated individuals are mapped to similar phenotypes. We consider this convenient for the evolutionary process.

The deletion and insertion procedures cause the variation of the number of genes (line segments), allowing the evolution of stencils with a different number of elements. Also, we implement a mechanism that boosts the mutation rate when the individuals in the population are too similar. This mechanism promotes diversity throughout the evolutionary process and prevents stagnation of the evolution.

Both variation operators preserve the validity of the stencils and the integrity of their line segments. We consider a stencil valid if all its line segments are different; all lines segments are located inside the limits of the grid; the number of line segments remains within preset ranges; all line segments have no null length; and no line segment contains another segment.

\subsection{Evaluation}\label{subsec_evaluation}

We implemented an automatic fitness assignment scheme to autonomously guide the evolution of stencils. In short, the fitness of a stencil consists in the evaluation of the best glyphs it can produce. Therefore, each stencil is evaluated in how well it performs in drawing glyphs for the target characters.

The evaluation of each stencil takes a couple of steps. As mentioned in the subsection~\ref{subsubsec_representation}, each stencil has several line segments that can be activated to express glyphs. First, the system chooses a character for which the stencil has to produce glyphs. Next, among the stencil's segments, we search which mask of active segments better expresses glyphs for the chosen character. This way, each mask stores the best use, or configuration, of the stencil found during the evaluation process to draw a given character. Each mask is evaluated by how the expressed glyph maximises a predetermined glyph evaluation function, which is later described in this section.
 
We use a hill-climbing algorithm to perform the search for the best stencil's configuration for each target character. 
The search starts with all the segments deactivated, activating one per step. At each step, all newly generated configurations are evaluated as a glyph for the target character. The search stops when no improvement in the evaluation is achieved, storing the best mask and the glyph's evaluation value. We also store the top $k$ solutions for each character as alternative configurations, although these do not enter in the fitness evaluation of the stencil. This process is repeated for all the target characters.

Each stencil's configuration is evaluated based on the visual similarity of its expression (see figure~\ref{fig_mapping}) with a preset glyph of the target character. This way, after the stencil's configuration and the target glyph are rendered to an image representation, the similarity between the two images is calculated using a pixel-by-pixel comparison using a \gls{RMSE} method.

\begin{equation}
    \text{\emph{fit\_exp\_1}}(s) = \sum_{g \in G, e_s \in E_s}{(1 - \text{RMSE}(e_s, g))/\mid G \mid}
    \label{formula_exp_1}
\end{equation}

\noindent Since each stencil is used to design multiple glyphs we created the fitness function \emph{fit\_exp\_1} to evaluate a given stencil. This function is defined by the equation~\ref{formula_exp_1}, where %
$s$ is the stencil to evaluate, %
$e_{s}$ is the expression of the best configuration found for the target character generated utilising stencil $s$, %
$E_s$ is the set of expressions of the best configurations generated utilising stencil $s$, %
$g$ is the preset glyph of the target character, %
, $G$ is the set of preset glyphs, $\mid G \mid$ cardinality of G. %
This function tends to $1$ if the stencil is able to generate glyphs similar to the target ones.

\section{Experiments}\label{sec_experiments}

\begin{figure}[!h]
    \centering
    \includegraphics[width=0.95\textwidth]{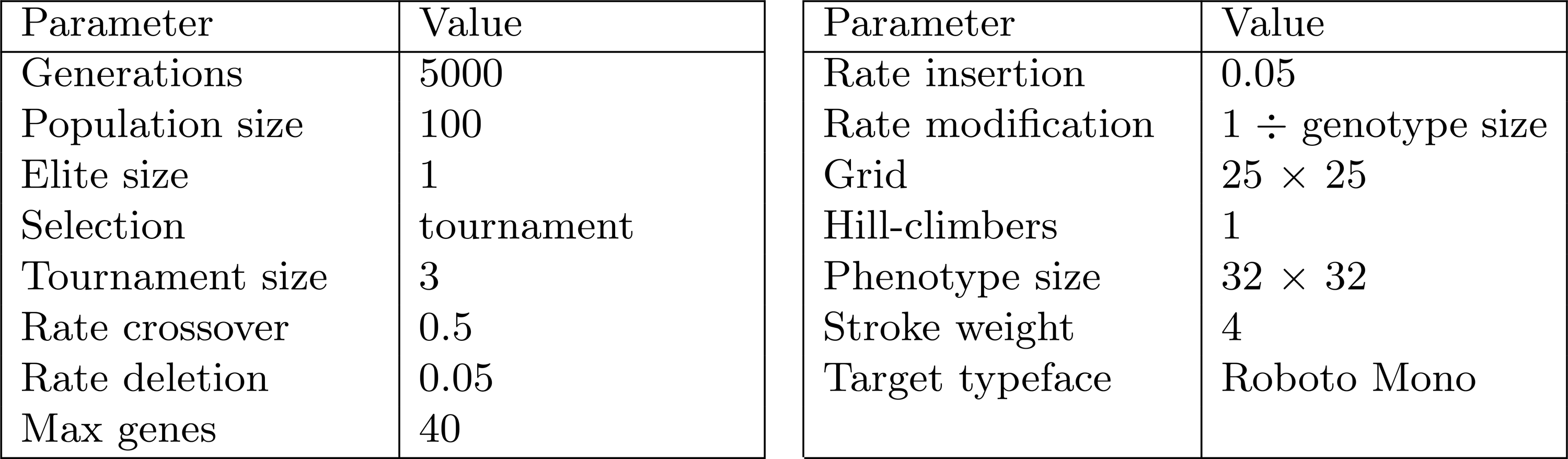}
    \caption{Experimental parameters}
    \label{table_exp_parameters}
\end{figure}

In this work, we conducted a series of experiments to explore and analyse the possibilities created with the system. The experimental parameters used in the experiments are summarised in table~\ref{table_exp_parameters}. We focus on the evolution of stencils to draw sans serif typefaces with glyphs for the uppercase (capital) letters of the Roman alphabet. 

\noindent Experiment~$1$ focuses on the ability of the system to evolve stencils (see subsection~\ref{subsec_exp_1}). Then, in experiment~$2$, we add two components to the stencils' evaluation that takes into consideration the number of elements and the gaps between them (see subsection~\ref{subsec_exp_2}). In experiment~$3$, we study the sharing of stencils' elements among the glyphs (see subsection~\ref{subsec_exp_3}). Finally, in experiment~$4$, we test the coherence provided by the evolved stencils by replacing their elements with different shapes (see subsection~\ref{subsec_exp_4}).

\subsection{Experiment $1$}\label{subsec_exp_1}

We begin our experiments by analysing the fitness values obtained. The convergence of the maximum fitness values over the generations is visualised in figure~\ref{fig_progression_fitness_1}. As one can see, the system is able to guide evolution and optimise the stencils' fitness. Regarding the rate of fitness convergence, high fitness values are attained in few generations.

In an early stage of the evolution, stencils tend to absorb the elements that overlap the area of the target glyph. After the maximum number of elements is reached, only adjustments to or replacements of existing elements are able to improve the fitness of the individuals. As a result, the fitness values increase faster during the initial stage than during the rest of the evolution.

One can observe that the maximum fitness values obtained (${\sim}0.785$) do not reach the theoretical maximum value ($1$). This behaviour may be related to: (i) the stencils have to satisfy many requirements specific to each character, which may reveal to be a difficult problem to solve; (ii) the maximum number of elements per stencil ($40$) is too low and perhaps more elements are needed to draw parts of the glyphs (see figure~\ref{fig_progression_num_elements_1}); (iii) the positioning resolution enabled by the grid is too low to allow some details of the glyphs to be represented; and (iv) the use of just line segments is not the best way to render curved parts that compose some glyphs.

\begin{figure}
    \centering
    \includegraphics[width=1\textwidth]{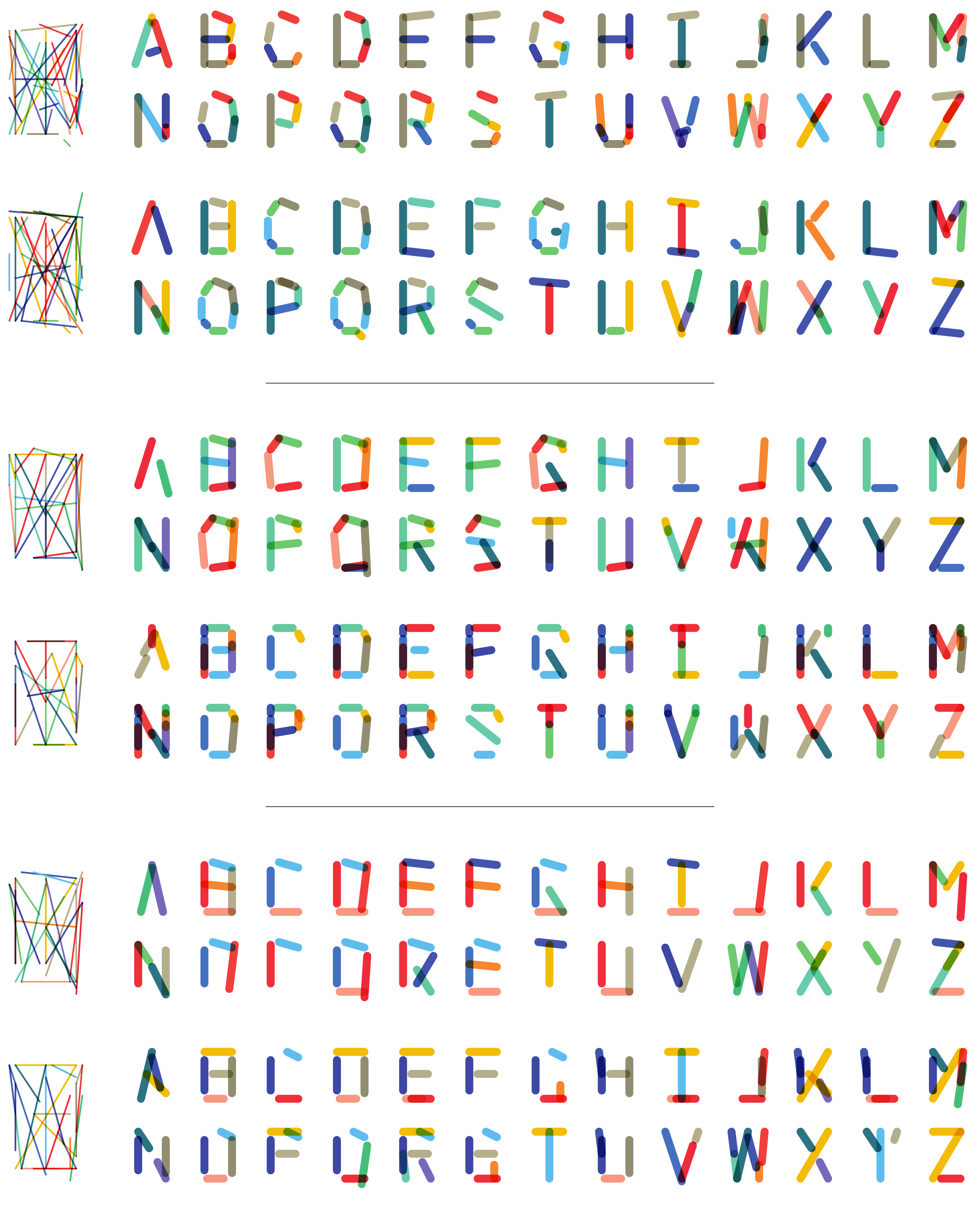}
    \caption{Typical results evolved in different runs in experiment~$1$ (top), experiment~$2$ (middle), and experiment~$3$ (bottom). To better identify each element of the stencils (left) in the corresponding glyphs (right), a random colour is used for each element. More results can be visualised at \urlVideoResults.}
    \label{fig_results}
\end{figure}

\noindent Figure~\ref{fig_results} (top) shows typical results evolved in different runs. More results can be visualised at \urlVideoResults. The results obtained demonstrate the ability of the system to evolve stencils that produce glyphs similar to the targets. Nevertheless, the system is able to produce diversity in terms of stencils, thus generating glyphs with different features. In addition, one can observe the reuse of stencils' elements among the glyphs. This behaviour is crucial as it creates coherence and unity in the glyphs.

It is possible to see many gaps between the segments that compose the glyphs. Furthermore, it seems that the representation adopted in this work strives for representing some parts of the glyphs, \emph{e.g.}, curves and other parts with a different thickness than the preset stroke weight of the stencils' elements. One possible solution for this issue could be the encoding of the stroke weight of each element in the respective gene to allow the evolution to modify it according to each situation. Regarding the appearance of the evolved stencils, most of them use the maximum number of elements, resulting in complex tangles of lines. Consequently, it is difficult to identify which elements of the stencil are used by each glyph.

\subsection{Experiment $2$}\label{subsec_exp_2}

In this experiment, we approach two issues identified in the results obtained in experiment~$1$: (i) the high number of elements that compose the stencils and (ii) the several gaps that appear between the elements. To do so, we designed the fitness function \emph{fit\_exp\_2}, which is a modification of \emph{fit\_exp\_1} (see equation~\ref{formula_exp_1}), to control the number of elements and minimise the gaps. The function \emph{fit\_exp\_2} is defined by the equation~\ref{formula_exp_2}, where $s$ is the stencil to evaluate, $size$ returns a value between $0.95$ and $1$, inversely proportional to the number of elements of $s$, and $gaps$ returns a value between $0.975$ and $1$, directly proportional to the percentage of elements' end points that are contained in other elements.

\begin{equation}
    \text{\emph{fit\_exp\_2}}(s) = \text{\emph{fit\_exp\_1}}(s) \times \text{\emph{reduce\_size}}(s) \times \text{\emph{reduce\_gaps}}(s)
    \label{formula_exp_2}
\end{equation}

\begin{figure}
    \centering
    \begin{minipage}{0.49\textwidth}
        \centering
        \includegraphics[width=\textwidth]{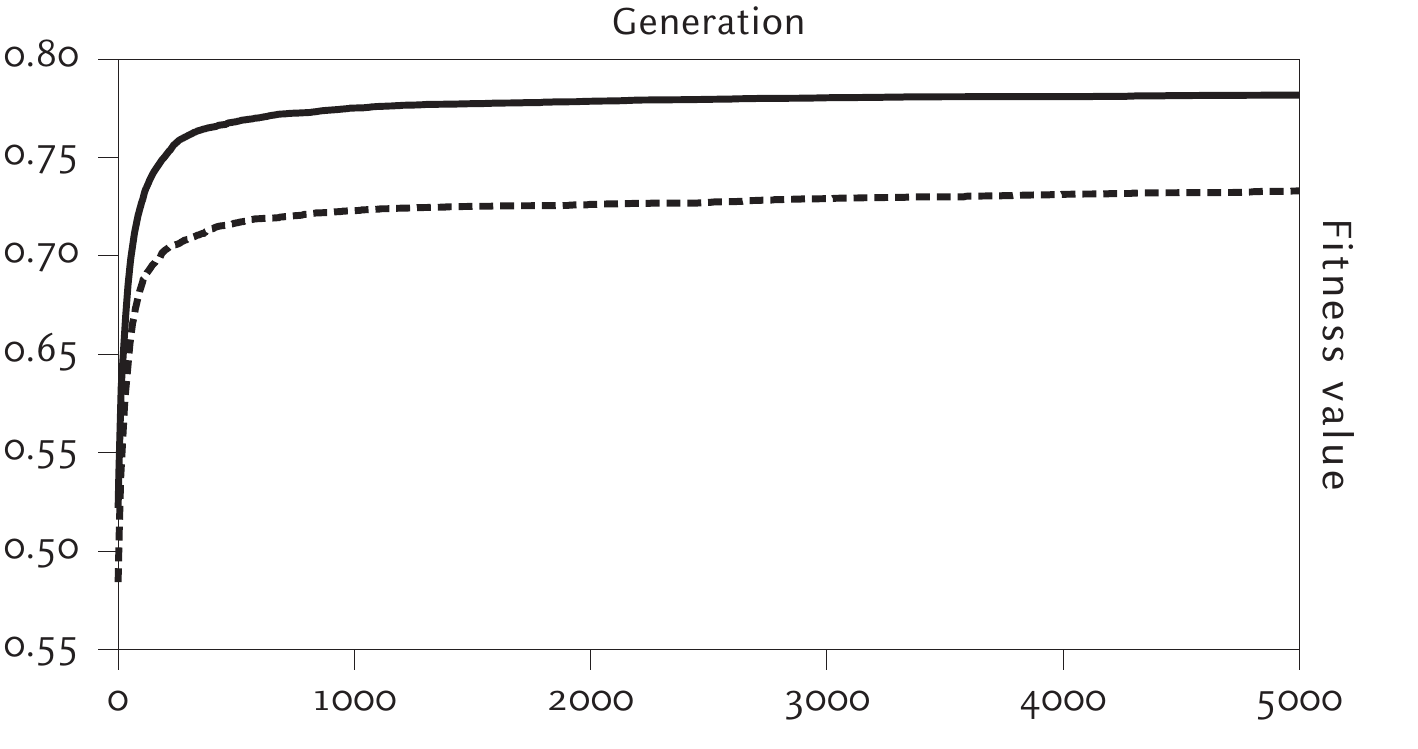}
        \caption{Progression of the fittest stencil's fitness over the generations when using the fitness functions \emph{fit\_exp\_1} (solid line) and \emph{fit\_exp\_2} (dashed line) to guide evolution. The visualised data is the average of $15$ runs.}
        \label{fig_progression_fitness_1}
    \end{minipage}\hfill
    \begin{minipage}{0.49\textwidth}
        \centering
        \includegraphics[width=\textwidth]{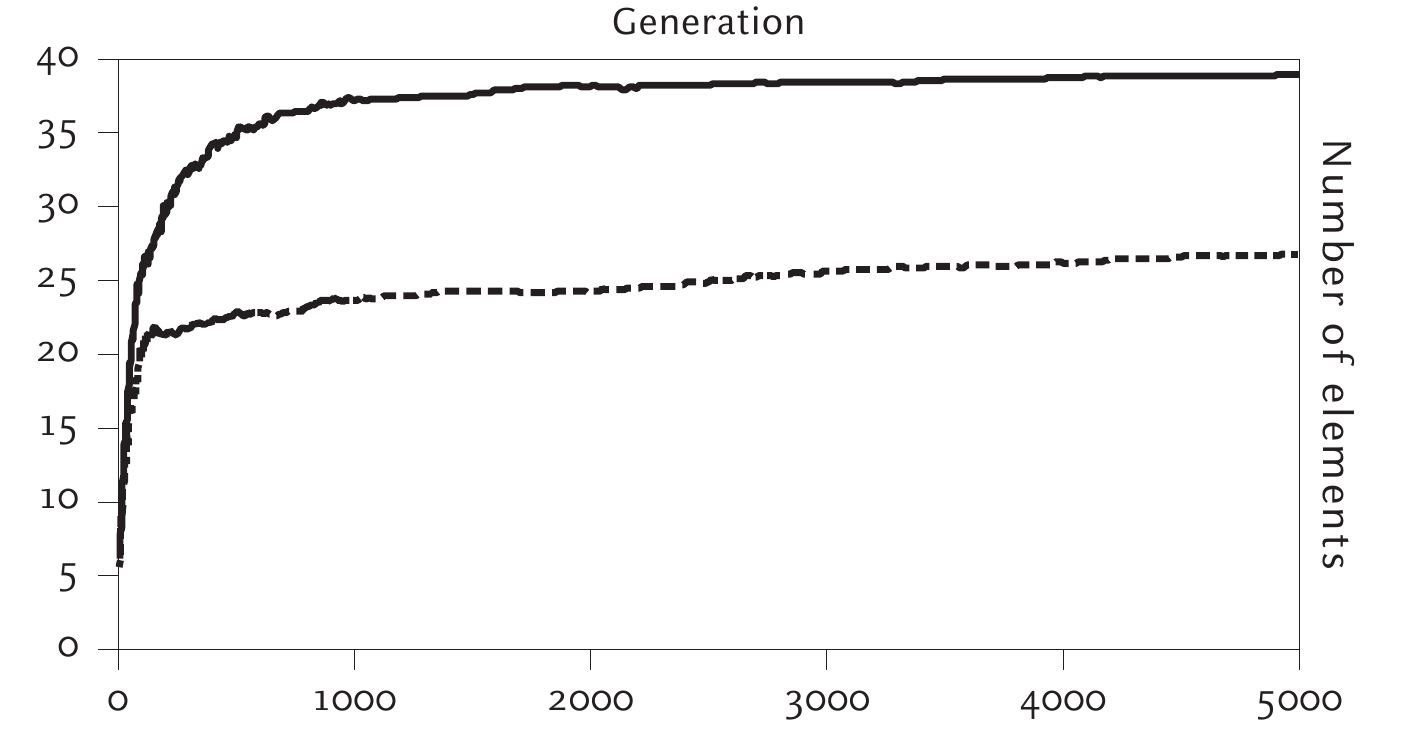}
        \caption{Progression of the number of elements used by the fittest stencils over the generations when using the fitness functions \emph{fit\_exp\_1} (solid line) and \emph{fit\_exp\_2} (dashed line) to guide evolution. The visualised data is the average of $15$ runs.}
        \label{fig_progression_num_elements_1}\label{fig_progression_shapes}
    \end{minipage}
\end{figure}

\noindent The convergence of the maximum fitness values over the generations is visualised in figure~\ref{fig_progression_fitness_1}. The behaviour of the fitness progression is similar to the one observed in experiment~$1$ but with a lower magnitude. When \emph{fit\_exp\_2} is guiding evolution, stencils do not achieve fitness values as high as experiment~$1$. This is related to the pressure that $size$ and $gaps$ have in the stencils' evaluation. In this experiment, stencils have to draw proper glyphs, \emph{i.e.}, maximise \emph{fit\_exp\_1}, while compromising the number of elements they use, \emph{i.e.}, maximising \emph{reduce\_size}, and compromising some details' reproduction capacity as a consequence of removing gaps between the elements, \emph{i.e.}, maximising \emph{reduce\_gaps}.

Figure~\ref{fig_progression_num_elements_1} compares for \emph{fit\_exp\_1} and \emph{fit\_exp\_2} the progression of the number of elements used by the fittest stencils over the generations. The progression rates regarding the number of elements and fitness are similar. However, the difference in the number of elements in the two approaches is clear. In average, when \emph{fit\_exp\_2} guides evolution, stencils use roughly two-thirds of the elements in comparison with when \emph{fit\_exp\_1} is guiding ($26.8$ versus $39$, respectively). 

Figure~\ref{fig_results} (middle) shows typical results evolved in different runs. With \emph{fit\_exp\_2} guiding evolution, the system continues to generate stencils that are able to produce glyphs similar to the targets. However, it is noticeable the differences at the visual level in comparison with the results obtained in experiment~$1$. The glyphs are built with a reduced number of elements. Regarding the gaps, the visual results demonstrate the capacity of \emph{fit\_exp\_2} in guiding the evolution towards stencils with elements whose end points are contained in other elements, creating connections between the different them. This is reflected in the appearance of the stencils, which are more clean and elementary than the ones evolved in experiment~$1$. For this reason, it is easier to identify which elements of the stencil are used by each glyph.

The simplicity of the glyphs produced in this experiment makes the reuse of the stencils' elements among the glyphs more noticeable. However, the reduced number of elements forces the evolutionary process to compromise between the abstraction and the complete glyphs' structures, resulting in glyphs with some missing parts. In most of the cases where parts of the glyphs were missing, we noticed that those parts are typically specific to one glyph and therefore no other glyphs contain them.

\subsection{Experiment $3$}\label{subsec_exp_3}

In this experiment, we further investigate one stencils' property that we consider fundamental: the reuse of elements to draw different glyphs. We began by visualising this feature in the results obtained in experiment~$2$.

\begin{figure}[!b]
    \centering
    \includegraphics[width=0.525\textwidth]{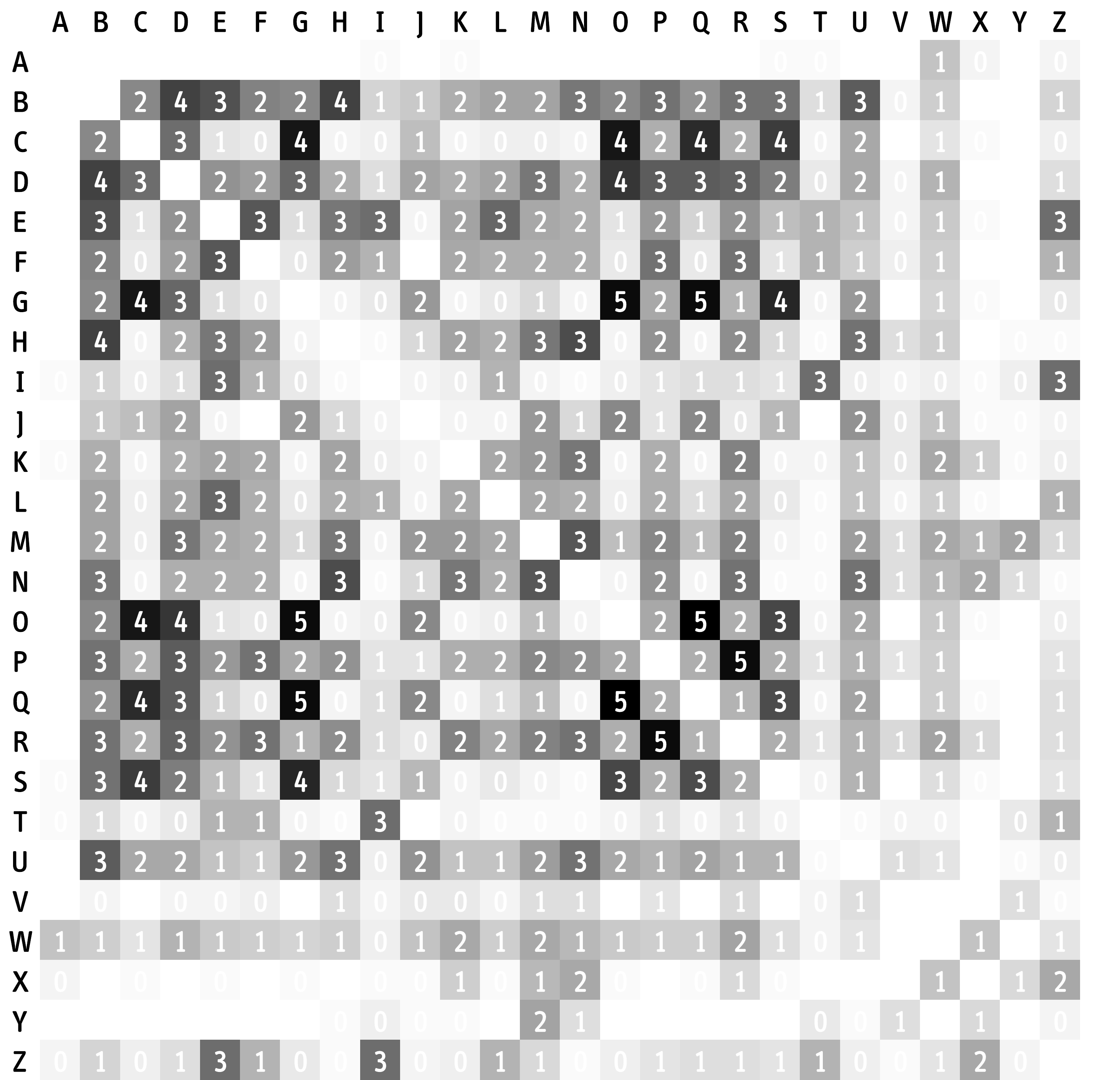}
    \caption{Visualisation of the number of elements shared by the different glyphs. The data corresponds to the average of the fittest stencils evolved in $15$ runs of experiment~$2$, which in average are composed of $26.8$ elements. Each value is rounded to its closest integer.} 
    \label{fig_shared_elements}
\end{figure}

Figure~\ref{fig_shared_elements} shows the average number of elements shared by the best glyphs produced with stencils evolved in experiment~$2$. The data demonstrates that the stencils reuse many of its elements to draw the different glyphs, predominantly the ones that are visually similar, \emph{e.g.}, %
\texttt{\lq{GO}\rq}, 
\texttt{\lq{GQ}\rq}, 
\texttt{\lq{OQ}\rq}, 
\texttt{\lq{PR}\rq}, 
\texttt{\lq{BD}\rq}, 
\texttt{\lq{BH}\rq}, 
\texttt{\lq{CG}\rq}, 
\texttt{\lq{CO}\rq}, 
\texttt{\lq{CQ}\rq}, 
\texttt{\lq{CS}\rq}, 
\texttt{\lq{DO}\rq}, 
and \texttt{\lq{GS}\rq}. 

Based on this analysis, we set one last evolutionary experiment: evolve stencils to draw all glyphs with only a subset being evaluated by the fitness function. This way, we designed the fitness function \emph{fit\_exp\_3}, which is based on \emph{fit\_exp\_2} (see equation~\ref{formula_exp_2}), to guide evolution in this experiment. The difference to \emph{fit\_exp\_2} is that instead of using the evaluation of all glyphs from the preset $G$, it uses a subset $L$, where $L \subseteq G$, $L = \left\{ {\texttt{\lq{B}\rq}, \texttt{\lq{I}\rq}, \texttt{\lq{Q}\rq}, \texttt{\lq{V}\rq}, \texttt{\lq{W}\rq}, \texttt{\lq{X}\rq}}\right\}$. We resorted to the data visualised in figure~\ref{fig_shared_elements} to test different combinations of glyphs for the set $L$ until we got a small set of glyphs that could provide several parts that could be used to draw most of the target glyphs.

\begin{figure}[!h]
    \centering
    \begin{minipage}{0.49\textwidth}
        \centering
        \includegraphics[width=1\textwidth]{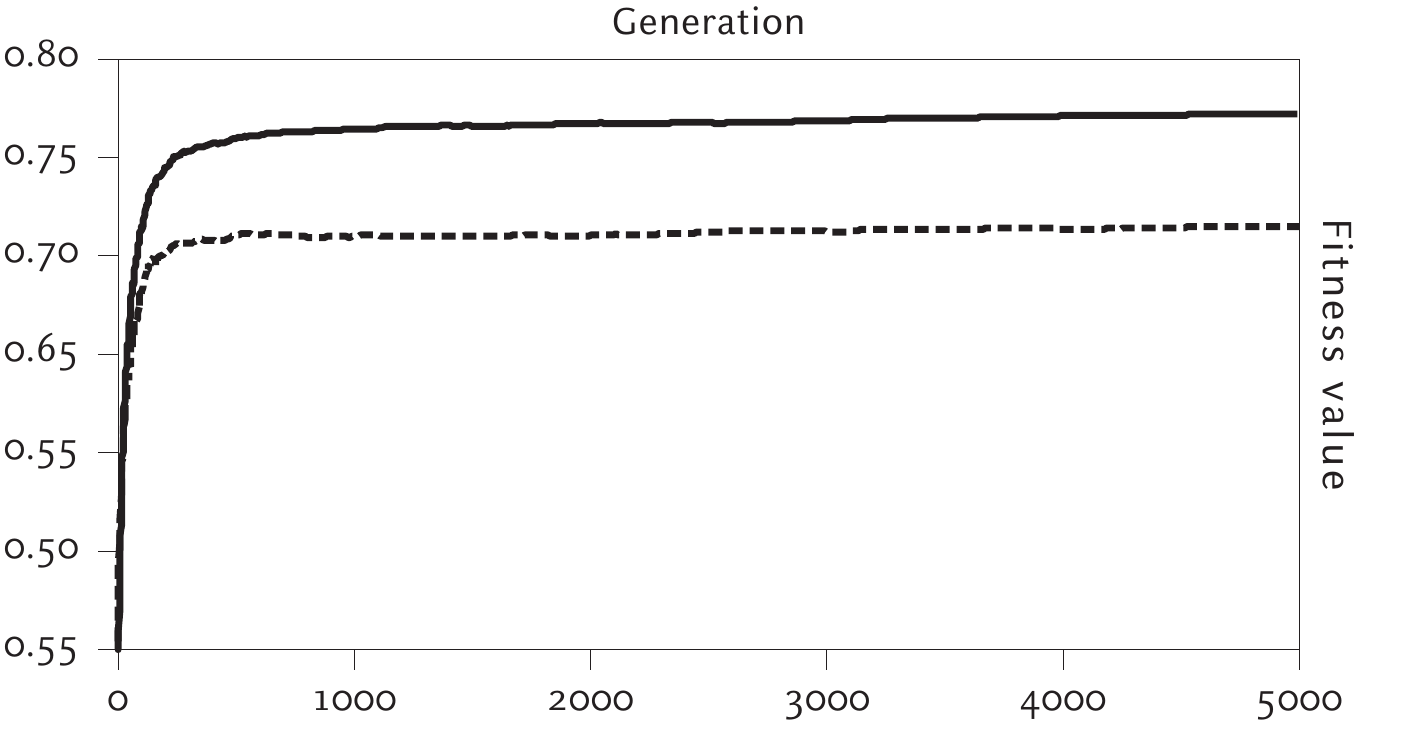}
        \caption{Progression of the average fitness of the glyphs that are being evaluated by the fitness function \emph{fit\_exp\_3} (solid line) and evolution of the average fitness of the remaining glyphs (dashed line) over the generations. The visualised data is the average of $15$ runs.}
        \label{fig_progression_fitness_2}
    \end{minipage}\hfill
    \begin{minipage}{0.49\textwidth}
        \vspace{-22pt}
        \centering
        \includegraphics[width=\textwidth]{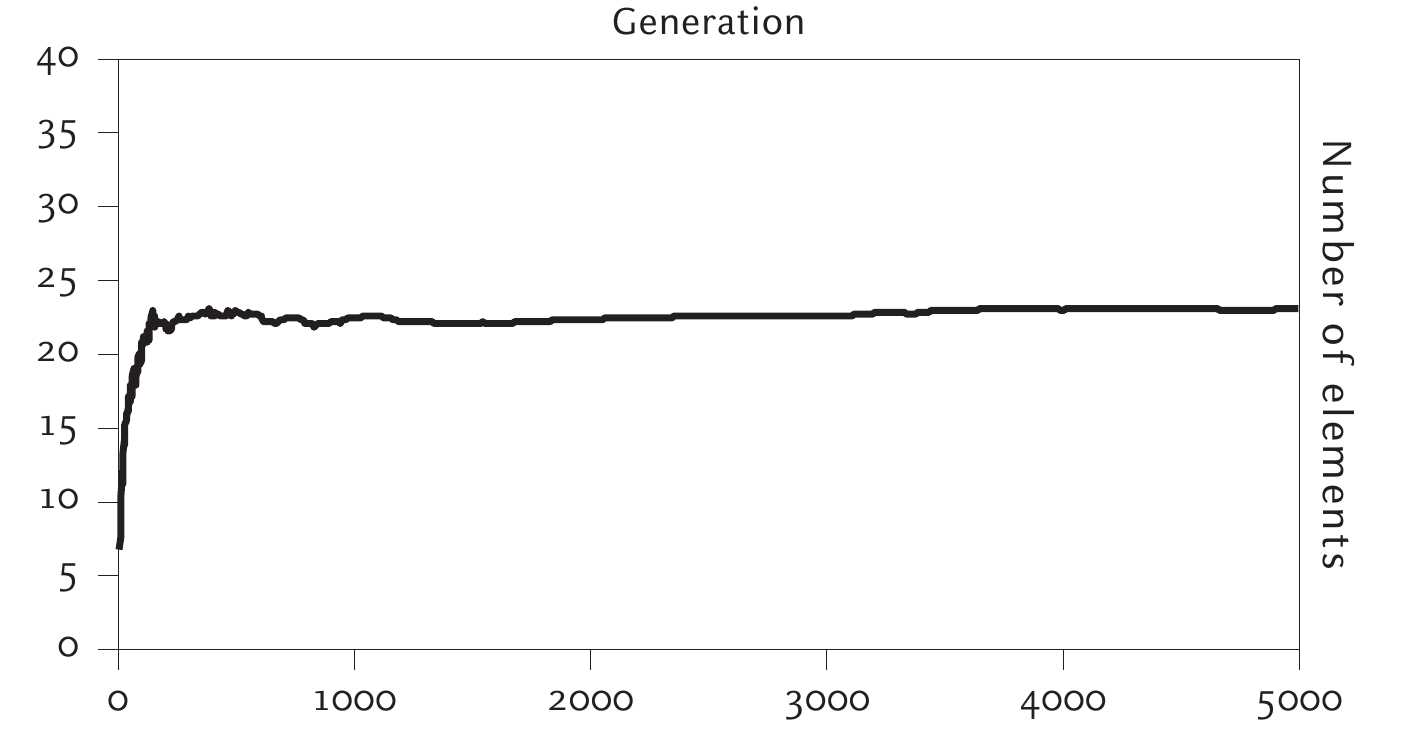} 
        \caption{Progression of the number of elements used by the fittest stencils over the generations when using the fitness function \emph{fit\_exp\_3} to guide evolution. The visualised data is the average of $15$ runs.}
        \label{fig_progression_num_elements_2}
    \end{minipage}
\end{figure}

The progression of the average fitness of the glyphs that are being evaluated by the fitness function, along with the progression of the average fitness of the remaining glyphs, is visualised in figure~\ref{fig_progression_fitness_2}. One can see that \emph{fit\_exp\_3} is able to improve the average fitness of all glyphs although only a small subset is considered by the fitness function. This demonstrates that stencils evolved to draw a small set of glyphs can be used to draw other glyphs.

Figure~\ref{fig_results} (bottom) shows typical results evolved in different runs. As one can see, the resulting glyphs use few elements and are quite simple. Nevertheless, just a few are hard to recognise due to some missing parts. It is noticeable the reuse of elements among the glyphs. Regarding the stencils, the number of elements used stabilises at $23$ (see figure~\ref{fig_progression_num_elements_2}), thus being more elementary than the ones evolved in experiment~$1$ ($39$~elements) and experiment~$2$ ($26.8$~elements).

\subsection{Experiment $4$}\label{subsec_exp_4}

In this last experiment, we explore the creative possibilities created with the stencils evolved with the system. To do so, we asked two graphic designers to create a series of typefaces using evolved stencils. We created a program that takes one stencil as input and replaces its elements with custom shapes designed by the user. The user is able to set which shape should replace each stencil's element, or opt for a random configuration. The program assembles the final glyphs on the fly. Therefore, the user can gain quickly an idea of what sort of effect any change has on the final glyphs.

\begin{figure}
    \centering
    \includegraphics[width=0.9\textwidth]{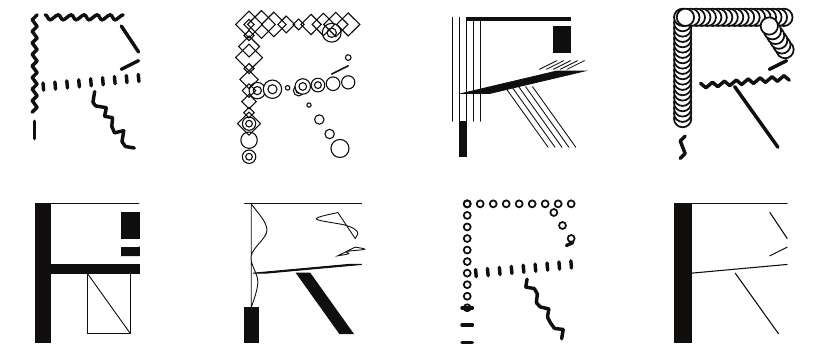}
    \caption{Elements of a stencil replaced with different shapes}
    \label{fig_stencil_application_r}
\end{figure}

\begin{figure}
    \centering
    \includegraphics[width=1\textwidth]{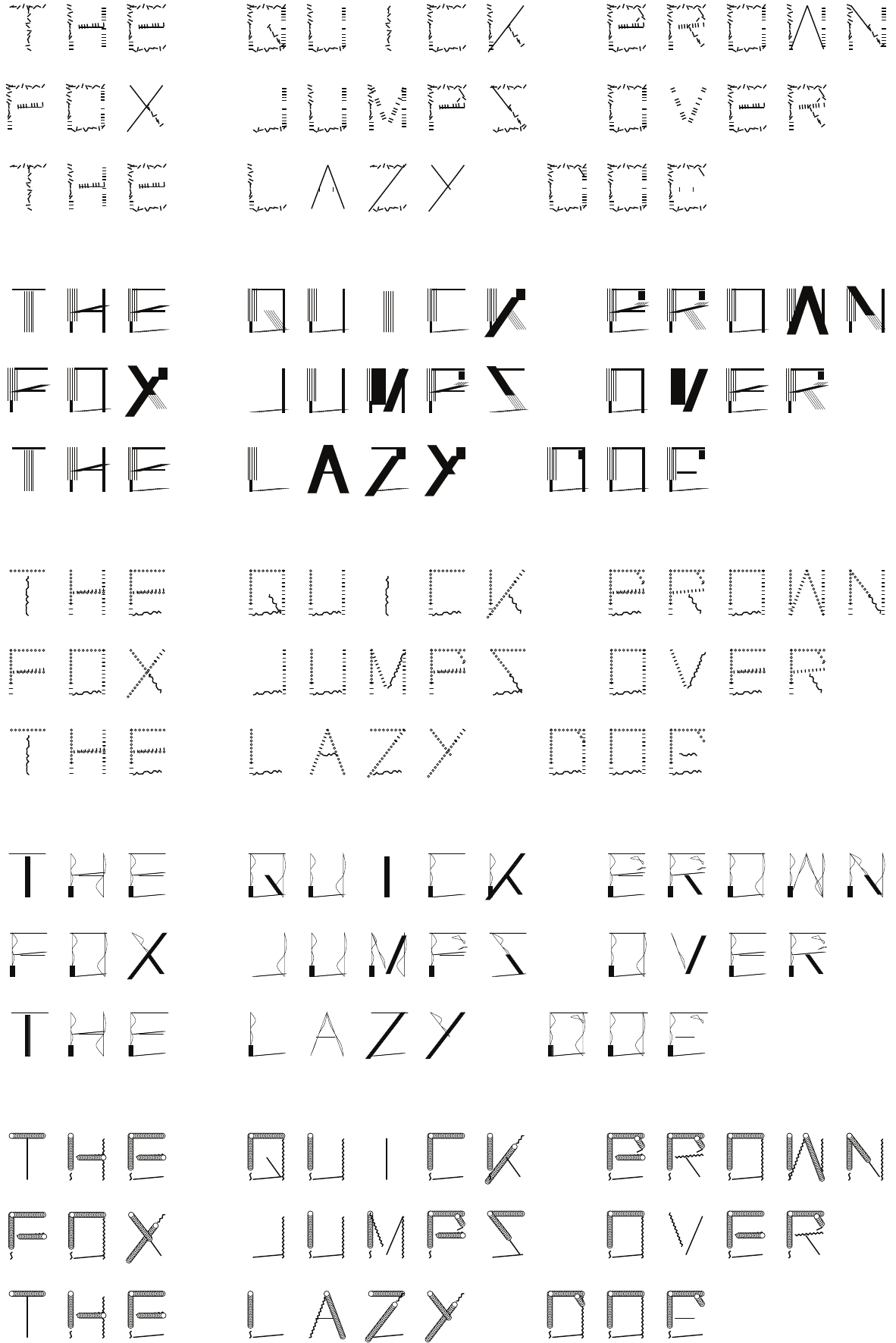}
    \caption{One evolved stencil with its elements replaced with different sets of shapes. More results can be visualised at \urlVideoResults.}
    \label{fig_stencil_application}
\end{figure}

\noindent As one can see in figure~\ref{fig_stencil_application_r}, both designers explored a wide range of shapes and strokes to give body to the final glyphs. Some are made up of dots, squares, lines, circles, or dashes; others consist of solid shapes or repetitions of random shapes with random sizes. However, as one can see in figure~\ref{fig_stencil_application}, the stencil is able to maintain coherence across the glyphs, thus showing a similar style. When the users employ randomness into their custom shapes, our approach is able to generate always changing glyphs that maintain coherence. More visual results can be visualised at \urlVideoResults.

The results demonstrate the ability of this approach to provide creative possibilities even within the fixed structure of the stencil. The basis structure of the stencil is maintained to ensure the legibility of the glyphs. However, a wide variety of expressive styles and shapes can be used to give body to the final glyphs. This way, one is able to produce expressive and functional typefaces, exploring different compromises between the expressiveness and the legibility of the glyphs while maintaining coherence.

The possibilities enabled by this approach are vast. One can evolve one stencil and then use it with different sets of shapes and strokes; or evolve different stencils and use them with the same set of shapes. In both approaches, coherence would still prevail among the final glyphs.

\section{Conclusions and Future Work}\label{sec_conclusions}

We have described and tested an evolutionary system for the generation of type stencils. A series of experiments were conducted to test and explore the system. We highlight the following contributions: %
(i) the evolution of stencils capable of producing legible, coherent glyphs; %
(ii) the reuse of the stencils' elements among the resulting glyphs and their consequent coherence and unity; %
(iii) the compromise between the simplicity of the evolved stencils and the drawing of several glyphs; %
(iv) the ability of stencils evolved to draw a small set of glyphs to draw other glyphs; %
(v) the creative possibilities created with the evolved stencils, even within their fixed structure; %
and (vi) the exploration of different compromises between the expressiveness and the legibility of the glyphs while maintaining coherence.

We do not expect our approach to compete with the more traditional type design approaches, or to replace the designer, or to output final design solutions. We see our work as an exploration of how new tools can be developed for the prototyping and play with the design of new type forms, regardless of the knowledge of the user in the domain.

We identify some applications for the type of results we have obtained. They could provide concepts and ideas to designers to design bespoke typefaces. The expressiveness of the results may produce eye-catching headlines for books, magazines, advertisements and therefore carry the message in an effective way. Also, the results could be used to create strong visual identities through the design of unique logotypes (see, \emph{e.g.}, the visual identity for Pick me Up 2016 and the identity for Tess Management). Also in the domain of visual identities, this approach could be used to design other visual elements other than letters, \emph{e.g.}, signage and symbols to display information and communicate messages. Finally, we consider the results may offer the wider audience a deeper understanding of the anatomy of type and the way we see letters.

Future work will focus on: %
(i) the exploration of other fitness assignment schemes to create more diversity among the evolved stencils; %
(ii) the use of an evolutionary approach to find other, and possibly smaller, subsets $L$ for experiment~$3$; %
(iii) the integration of other graphic primitives as the elements of the stencils, including curves and more complex shapes; %
(iv) the experimentation with \gls{IEC} to evolve the shapes and settings for the stencils' elements; %
(v) the implementation of routines, automatic and/or interactive, to fine-tune the evolved stencils that would update the final glyphs in a live fashion, \emph{e.g.}, merge of similar and redundant elements, clustering of elements' coordinates, and edition of the elements; %
and (vi) the creation of real stencils based on evolved stencils.

\section{Acknowledgements}

This research is partially funded by Funda\c{c}\~ao para a Ci\^encia e Tecnologia (FCT), Portugal, under the grants SFRH/BD/90968/2012 and SFRH/BD/105506/2014; %
and is based upon work from COST Action CA15140: ImAppNIO, supported by COST (European Cooperation in Science and Technology): \href{www.cost.eu}{www.cost.eu}. %
We would also like to express our gratitude to NVIDIA for providing us one Titan Xp GPU to support our research.

\bibliographystyle{splncs03} %
\bibliography{evotype3}

\begin{thebibliography}{10}
\providecommand{\url}[1]{\texttt{#1}}
\providecommand{\urlprefix}{URL }

\bibitem{lupton2004a}
Lupton, E.: Thinking with Type: A Critical Guide for Designers, Writers,
  Editors, and Students. Princeton Architectural Press, 1st edn. (2004)

\bibitem{martins2015a}
Martins, T., Correia, J., Costa, E., Machado, P.: Evotype: Evolutionary type
  design. In: Johnson, C., Carballal, A., Correia, J. (eds.) Evolutionary and
  Biologically Inspired Music, Sound, Art and Design, Lecture Notes in Computer
  Science, vol. 9027, pp. 136--147. Springer International Publishing (2015),
  \url{http://dx.doi.org/10.1007/978-3-319-16498-4\_13}

\bibitem{martins2016a}
Martins, T., Correia, J.a., Costa, E., Machado, P.: Evotype: From shapes to
  glyphs. In: Proceedings of the Genetic and Evolutionary Computation
  Conference 2016. pp. 261--268. GECCO '16, ACM, New York, NY, USA (2016),
  \url{http://doi.acm.org/10.1145/2908812.2908907}

\bibitem{mrm07a}
Machado, P., Romero, J., Manaris, B.: Experiments in computational aesthetics:
  An iterative approach to stylistic change in evolutionary art. In: Romero,
  J., Machado, P. (eds.) The Art of Artificial Evolution: A Handbook on
  Evolutionary Art and Music, pp. 381--415. Springer Berlin Heidelberg (2007)

\bibitem{butterfield2000a}
Butterfield, I., Lewis, M.: Evolving fonts (2000), consulted in
  \url{http://accad.osu.edu/~mlewis/AED/Fonts/} on November 2017

\bibitem{lund2000a}
Lund, A.: Evolving the shape of things to come: A comparison of direct
  manipulation and interactive evolutionary design. In: International
  Conference on Generative Art. Domus Argenia, Rome, Italy (2000)

\bibitem{levin2001a}
Levin, G., Feinberg, J., Curtis, C.: Alphabet synthesis machine (2001),
  consulted in \url{http://www.flong.com/projects/alphabet/} on November 2017

\bibitem{unemi2003a}
Unemi, T., Soda, M.: An iec-based support system for font design. In:
  Proceedings of the {IEEE} International Conference on Systems, Man {\&}
  Cybernetics: Washington, D.C., USA, 5--8 October 2003. pp. 968--973 (2003)

\bibitem{schmitz2004a}
Schmitz, M.: genotyp, an experiment about genetic typography. Presented at
  Generative Art Conference 2004 (2004)

\bibitem{kuzma2008a}
Kuzma, M.: Interactive Evolution of Fonts. Master's thesis, Technical
  University of Ko{\v s}ice (2008)

\bibitem{yoshida2010a}
Yoshida, K., Nakagawa, Y., K{\"o}ppen, M.: Interactive genetic algorithm for
  font generation system. In: World Automation Congress, 2010. pp. 1--6. TSI
  Press. (2010)

\bibitem{kindel2007a}
Kindel, E.: The `plaque d{\'e}coup{\'e}e universelle': a geometric sanserif in
  1870s paris. In: Typography papers 7, pp. 71--80. Hyphen Press, The
  Department of Typography \& Graphic Communication, University of Reading
  (2007)

\bibitem{craig2012a}
Craig, J., Scala, I.K., Bevington, W.: Designing with Type: The Essential Guide
  to Typography. Watson-Guptill Publications, 5th edn. (2006)

\bibitem{eiben2015a}
Eiben, A.E., Smith, J.E.: Introduction to Evolutionary Computing. Natural
  computing series, Springer, Berlin, Heidelberg, Paris (2015)

\end{thebibliography}
\end{document}